# Boosting Theory-of-Mind Performance in Large Language Models via Prompting

**Shima Rahimi Moghaddam\*, Christopher J. Honey**
Johns Hopkins University, Baltimore, MD, USA.

\* Correspondence to: sh.rahimi.m@gmail.com

## Abstract

Large language models (LLMs) excel in many tasks in 2023, but they still face challenges in complex reasoning. Theory-of-mind (ToM) tasks, which require understanding agents' beliefs, goals, and mental states, are essential for common-sense reasoning involving humans, making it crucial to enhance LLM performance in this area. This study measures the ToM performance of GPT-4 and three GPT-3.5 variants (Davinci-2, Davinci-3, GPT-3.5-Turbo), and investigates the effectiveness of in-context learning in improving their ToM comprehension. We evaluated prompts featuring two-shot chain of thought reasoning and step-by-step thinking instructions. We found that LLMs trained with Reinforcement Learning from Human Feedback (RLHF) (all models excluding Davinci-2) improved their ToM accuracy via in-context learning. GPT-4 performed best in zero-shot settings, reaching nearly 80% ToM accuracy, but still fell short of the 87% human accuracy on the test set. However, when supplied with prompts for in-context learning, all RLHF-trained LLMs exceeded 80% ToM accuracy, with GPT-4 reaching 100%. These results demonstrate that appropriate prompting enhances LLM ToM reasoning, and they underscore the context-dependent nature of LLM cognitive capacities.

## Introduction

What kinds of reasoning can large language models (LLMs) perform about everyday scenarios? Large language models (LLMs) have shown great success in a variety of tasks; however, they still struggle with tasks that require reasoning (Mahowald et al., 2023; Rae et al., 2021). One area of specific interest the is so-called "theory of mind" (ToM) reasoning, which involves tracking the mental state of agents, such as their goals, and what they know (Kosinski, 2023; Langley et al., 2022). Language models have dramatically advanced in the range of everyday questions to which they can accurately respond, but their ToM performance is thought to be relatively poor (Cuzzolin et al., 2020; Sap et al., 2022; Ullman, 2023). Here, we test the hypothesis that appropriate prompting can enhance the ToM performance of LLMs.

The capacity of LLMs to reliably perform ToM reasoning is important for several reasons. First, ToM is an essential element of social understanding, allowing people to participate in intricate social exchanges and to anticipate the actions or responses of others (Bedny et al., 2009; Heyes and Frith, 2014; Kidd and Castano, 2013; Moran et al., 2011; Seyfarth and Cheney, 2013; Young et al., 2007). Second, ToM is considered a complex cognitive capacity which is most highly developed in humans, and a small number of other animals (Krupenye and Call, 2019; Povinelli and Preuss, 1995). This may be because ToM relies on structured relational knowledge (e.g. agents can have goals; and agent X has goal G, but agent Y does not know that agent X has goal G). Models that work with social information and with humans will benefit from being able to reason about the mental states and beliefs of agents. Finally, ToM tasks often involve inferential reasoning. For instance, for successful ToM performance, LLMs need to reason based on unobservable information (e.g. hidden mental states of agents) that must be inferred from context rather than parsed from the surface text (e.g. explicitly stated features of a situation). Hence, assessing and improving these models' proficiency in ToM tasks could offer valuable insights into their potential for a wider range of tasks that require inferential reasoning.

In-context learning approaches can enhance the reasoning capacity of LLMs. Brown et al. (Brown et al., 2020) showed that, for sufficiently large language models (+100B parameters), one can enhance models' performance using only few-shot task demonstrations that are specified only through the input to the model (i.e. at inference time, without weight updates). This form of performance boosting is usually referred to as "few-shot learning". Wei et al. (Wei et al., 2022) later showed that the ability of LLMs to perform complex reasoning was improved when the few-shot examples in the prompt contain the reasoning steps for reaching a conclusion ("chain-of-thought reasoning") (Magister et al., 2022). Moreover, Kojima et al. (Kojima et al., 2022) showed that, even in the absence of exemplar demonstrations, instructing language models to think "step-by-step" enhances their reasoning performance. There is not currently a theoretical understanding of why these prompting techniques are beneficial, however some recent studies have explored the effects of compositional structure and local dependencies in training data on efficacy of these methods (Hahn and Goyal, 2023; Prystawski and Goodman, 2023).

The capability of LLMs to perform ToM reasoning is supported by some studies (Bubeck et al., 2023; Kosinski, 2023), but questioned by others (Sap et al., 2022; Trott et al., 2022; Ullman, 2023). Though this prior literature provides many insights into ToM in LLMs, the quantitative evaluations of ToM performance have two main limitations. First, they examine LLMs' ToM performance only on single-word or multiple-option completion (Kosinski, 2023; Sap et al., 2022; Trott et al., 2022; Ullman, 2023). However, LLMs may benefit from freely producing answers with multiple parts and speculating over multiple possibilities, rather than being assessed on a single word completion. Second, most of the work criticizing the ToM abilities of LLMs relied on either zero-shot testing (Trott et al., 2022; Ullman, 2023) or provided examples that lacked step-by-step reasoning toward an answer (Sap et al., 2022). Yet, the type of output generated by LLMs can be highly context-sensitive (Sejnowski, 2023). Therefore, we asked whether recent LLMs might exhibit improved ToM performance when provided with suitable prompts.

Here we evaluate the performance of LLMs faced with ToM comprehension questions and we explore whether this performance can be boosted using prompting methods such as step-by-step thinking, few-shot learning, and chain-of-thought reasoning (Brown et al., 2020; Kojima et al., 2022; Wei et al., 2022). Improving inferential reasoning performance by prompting is important because it is a flexible approach that does not require additional training or large new datasets. Further, if effective prompting techniques guide LLMs towards generating higher-quality ToM responses, this contributes to the overall reliability of their reasoning in wide-ranging everyday applications.

# Methods

## Models

We studied the four most recent GPT models from the Open AI family. These were GPT-4 (OpenAI, 2023a) as well as the Davinci-2, Davinci-3, and GPT-3.5-Turbo models, which are considered GPT-3.5 variants that improve on GPT-3 (Brown et al., 2020; Ouyang et al., 2022). These are all large models (+100B parameters), but they differ in their training methods (OpenAI, 2023c). Davinci-2 (API name: text-davinci-002) is one of the GPT-3.5 models which (in addition to the GPT-3 curricula) was also trained with supervised fine-tuning on human-written demonstrations (OpenAI, 2023b; Stiennon et al., 2020). Davinci-3 (API name: text-davinci-003), another GPT-3.5 model, is an upgraded version of Davinci-2 which was further trained with Reinforcement Learning from Human Feedback (RLHF) using Proximal Policy Optimization (OpenAI, 2023b; Ouyang et al., 2022; Stiennon et al., 2020). GPT-3.5-Turbo (the original version of ChatGPT) (OpenAI, 2023b) is yet another GPT-3.5 model, trained with both fine-tuning on human-written demonstrations and RLHF, then further optimized for conversation. GPT-4 is the most recent GPT model as of April 2023 (OpenAI, 2023a); there are few published details of the size and training methods for GPT-4, however, it appears to have undergone more intensive training with RLHF for better alignment with human intention (OpenAI, 2023a). We tested all models in a setting with temperature equal to 0.4 and the maximum length of generated text set to 150 tokens.

## Experimental Design

To examine the comprehension ability of these models on ToM scenarios, we evaluated their comprehension accuracy on both ToM scenarios and Control scenarios. The Control scenarios describe a scene ("Photo") without any agents. We refer to them as Photo scenarios (Supplement A). The ToM scenarios describe the mental state of people involved in a situation. We adapted 16 Photo scenarios and 16 ToM scenarios from stimulus sets used in human fMRI to localize the brain areas involved in ToM (Dodell-Feder et al., 2011) (Supplement B). These scenarios match in their general difficulty; however, they differ in the need to reason based on individuals' state of mind in the scenario. Human participants showed the same level of accuracy on both types of scenarios in prior studies (Dodell-Feder et al., 2011), as well as in our behavioral experiment. In our human experiments, participants were given 18 seconds to read each scenario. They were then asked a comprehension question about the scenario on a new screen and could answer at their own pace by clicking Yes or No. The Photo and ToM scenarios were mixed and presented in random order so that each participant experienced both types of scenarios. Participants showed similar accuracy across Photo (86% ±4%) and ToM (87% ±4%) scenarios (see Supplement C).

**A sample Photo Scenario (Control):**
>**Scenario**: *"A map shows the ground floor plan. A photocopy was sent to the architect yesterday, but at the time the kitchen door was missing. It was added to the map this morning."*
>**Question**: Does the architect's photocopy show the door?

**A sample ToM Scenario:**
>**Scenario:** *"The morning of the high school dance Sarah placed her high heel shoes under her dress and then went shopping. That afternoon, her sister borrowed the shoes and later put them under Sarah's bed."*
>**Question:** When Sarah gets ready, does she assume her shoes are under her dress?

## Dependent Variable

We measured the models' accuracy in answering comprehension questions following the brief scenario. Since LLMs are probabilistic models, we ran each prompt 20 times for every model, restarting the models for every repetition to ensure that the performance history of the models was not included, and to measure the reliability of the model's answers across re-initialization. Accuracy was defined as the proportion of correct responses across 320 questions (16 scenarios x 20 repetitions).

To evaluate the performance of each model, we analyzed their response to each prompt. To measure the accuracy of each model, we manually checked all the LLM-generated responses and labeled them as correct or incorrect. Due to the complexity of natural language, it was not always possible to label model output as correct or incorrect based on match to a template answer (e.g. "Yes" or "No"). We labeled an answer as correct if it had the correct conclusion (see Examples 1-5 in "Correct responses" below) or if the model included the correct reasoning as one possibility amongst multiple reasonable responses (see Example 6 in "Correct responses" below). We labeled responses as incorrect if they arrived at an incorrect overall conclusion (see Examples 1-4 in "Incorrect responses" below) or if the responses were inconclusive (see Examples 5 and 6 in "Incorrect responses" below). Below are the examples of correct and incorrect responses to a ToM comprehension question:

**Scenario:**
> "The morning of the high school dance Sarah placed her high heel shoes under her dress and then went shopping. That afternoon, her sister borrowed the shoes and later put them under Sarah's bed."

**Question:**
> When Sarah gets ready, does she assume her shoes are under her dress?

**Correct Responses:**

- Example 1: Sarah placed her shoes under her dress before she went shopping, but her sister borrowed them and put them under Sarah's bed. Sarah doesn't know that her sister borrowed her shoes, so she may assume that they are still under her dress.

- Example 2: Sarah placed her high heel shoes under her dress before going shopping, but her sister borrowed the shoes and put them under Sarah's bed later. So, Sarah doesn't know that her sister borrowed and returned the shoes. Therefore, the answer is: Yes, she assumes her shoes are under her dress.

- Example 3: Sarah may assume her shoes are still under her dress because that's where she placed them in the morning.

- Example 4: Sarah placed her high heel shoes under her dress before she went shopping, so she knows where she left them. Her sister borrowed the shoes and later put them under Sarah's bed, but Sarah doesn't know that yet. So, the answer is: Yes, Sarah assumes her shoes are under her dress unless she discovers otherwise.

- Example 5: In the morning, Sarah placed her high heel shoes under her dress. Later, her sister borrowed the shoes and put them under Sarah's bed. Therefore, when Sarah gets ready, she may assume her shoes are still under her dress, but they are actually under her bed.

- Example 6: Based on this scenario, it is unclear whether Sarah assumes her shoes are under her dress when she gets ready. It depends on whether her sister told her she borrowed them and put them under the bed.

**Incorrect Responses:**

- Example 1: Sarah placed her shoes under her dress in the morning and then went shopping. Her sister borrowed the shoes and put them under Sarah's bed. So, Sarah doesn't know that her shoes are under her bed. So, the answer is: No, Sarah doesn't assume her shoes are under her dress.

- Example 2: No, Sarah does not assume her shoes are under her dress.

- Example 3: First, Sarah placed her high heel shoes under her dress in the morning. Then, her sister borrowed the shoes and later put them under Sarah's bed. Therefore, Sarah would not assume her shoes are under her dress as they were moved by her sister.

- Example 4: In the morning, Sarah placed her high heel shoes under her dress, but then went shopping. This means that when she returned home, she did not check if her shoes were still under her dress. Later, her sister borrowed the shoes and put them under Sarah's bed. Therefore, Sarah cannot assume that her shoes are under her dress as they were moved by her sister.

- Example 5: It does not specify in the scenario whether or not Sarah assumes her shoes are under her dress.

- Example 6: Sarah placed her high heel shoes under her dress in the morning. She then went shopping. Her sister later borrowed the shoes and put them under Sarah's bed. Based on this information, it is unclear if Sarah assumes her shoes are still under her dress when she gets ready for the dance.

## Experimental Conditions

To measure the efficacy of in-context learning (ICL) on ToM performance, we examined each scenario with four types of prompting: (1) Zero-shot (no ICL); (2) Zero-shot with step-by-step (SS) thinking; (3) Two-shot chain-of-thought (CoT) reasoning; (4) Two-shot CoT reasoning with SS thinking. Below are the examples of each prompting method (Figure 1).

**Zero-Shot**

**Prompt:**
Read the scenario and answer the following question:

Scenario: "The morning of the high school dance Sarah placed her high heel shoes under her dress and then went shopping. That afternoon, her sister borrowed the shoes and later put them under Sarah's bed."
Question: When Sarah gets ready, does she assume her shoes are under her dress?
A:

**Zero-Shot + Step-by-Step Thinking**

**Prompt:**
Read the scenario and answer the following question:

Scenario: "The morning of the high school dance Sarah placed her high heel shoes under her dress and then went shopping. That afternoon, her sister borrowed the shoes and later put them under Sarah's bed."
Question: When Sarah gets ready, does she assume her shoes are under her dress?
A: Let's think step by step:

**Two-Shot Chain of Thought Reasoning**

**Prompt:**
Read the scenario and answer the following question:

Scenario: "Anne made lasagna in the blue dish. After Anne left, Ian came home and ate the lasagna. Then he filled the blue dish with spaghetti and replaced it in the fridge."
Q: Does Anne think the blue dish contains spaghetti?
A: When Anne left the blue dish contained lasagna. Ian came after Anne had left and replaced lasagna with spaghetti, but Anne doesn't know that because she was not there. So, the answer is: No, she doesn't think the blue dish contains spaghetti.

Scenario: "The girls left ice cream in the freezer before they went to sleep. Over night the power to the kitchen was cut and the ice cream melted."
Q: When they get up, do the girls believe the ice cream is melted?
A: The girls put the ice cream in the freezer and went to sleep. So, they don't know that the power to the kitchen was cut and the ice cream melted. So, the answer is: No, the girls don't believe the ice cream is melted.

Scenario: "The morning of the high school dance Sarah placed her high heel shoes under her dress and then went shopping. That afternoon, her sister borrowed the shoes and later put them under Sarah's bed."
Question: When Sarah gets ready, does she assume her shoes are under her dress?
A:

**Two-Shot Chain of Thought Reasoning + Step-by-Step Thinking**

**Prompt:**
Read the scenario and answer the following question:

Scenario: "Anne made lasagna in the blue dish. After Anne left, Ian came home and ate the lasagna. Then he filled the blue dish with spaghetti and replaced it in the fridge."
Q: Does Anne think the blue dish contains spaghetti?
A: Let's think step by step: When Anne left the blue dish contained lasagna. Ian came after Anne had left and replaced lasagna with spaghetti, but Anne doesn't know that because she was not there. So, the answer is: No, she doesn't think the blue dish contains spaghetti.

Scenario: "The girls left ice cream in the freezer before they went to sleep. Over night the power to the kitchen was cut and the ice cream melted."
Q: When they get up, do the girls believe the ice cream is melted?
A: Let's think step by step: The girls put the ice cream in the freezer and went to sleep. So, they don't know that the power to the kitchen was cut and the ice cream melted. So, the answer is: No, the girls don't believe the ice cream is melted.

Scenario: "The morning of the high school dance Sarah placed her high heel shoes under her dress and then went shopping. That afternoon, her sister borrowed the shoes and later put them under Sarah's bed."
Question: When Sarah gets ready, does she assume her shoes are under her dress?
A: Let's think step by step:

**Figure 1. Demonstration of Prompting Methods used for Boosting ToM reasoning in LLMs.** Examples of 4 prompting types used to test the ToM performance of LLMs. Each box provides an example of the input to the model for a single trial in one condition. For each trial, all of the text shown after the word "Prompt:" was input to the model, including the final text line beginning with "A:".

# Results

## Zero-shot Performance

We first compared the models' zero-shot performance on Photo and ToM scenarios. We found that accuracy in Photo scenarios gradually increased with the recency of the models, with lowest performance in Davinci-2 and highest in GPT-4 (Figure 2 A). However, in contrast to Photo comprehension, the accuracy on ToM questions did not monotonically improve with the recency of the models: Davinci-2 was more accurate than Davinci-3, which was in turn more accurate than GPT-3.5-Turbo (Figure 2 B). Although the lower zero-shot ToM accuracy in GPT-3.5-Turbo in comparison to Davinci-2 may seem to imply that the Turbo model's reasoning performance is inferior, the primary reason for its lower accuracy was its tendency to provide an inconclusive response. Specifically, it would often state that there was insufficient information to determine the answer to the question (see examples 5-6 in "Incorrect responses" in Dependent Variable). However, more recent models were not always more equivocal in ToM responding: GPT-4 demonstrated a significantly greater ToM accuracy than all other models (independent t-test between GPT-4 and Davinci-2: $\Delta Acc = 0.11$, p-value <0.001) (Figure 2B). Overall, GPT-4 showed the best zero-shot performance in both Photo and ToM scenarios (Figure 2 A and B).

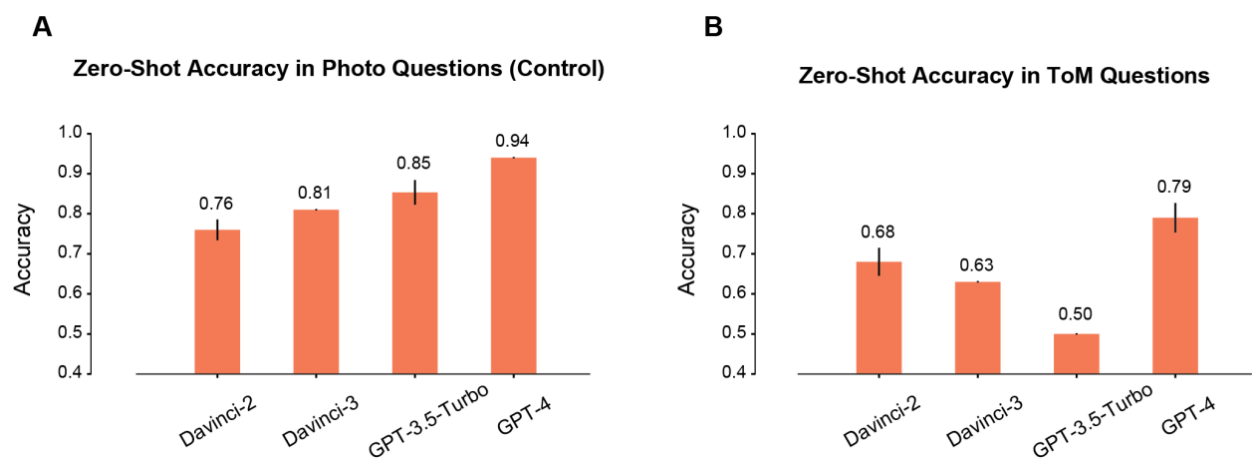

**Figure 2. Zero-Shot Accuracy of LLMs in Photo (Control) and ToM Comprehension Questions**. A) Zero-shot accuracy of LLMs in Photo scenarios (non-agentive) comprehension questions. B) Zero-shot accuracy of LLMs in ToM comprehension questions. The values show the mean accuracy for each model averaged over 320 measurements, as there are 16 scenarios of each type, and each scenario was tested 20 times with re-initialization. For plotting the error bars, we treated each "repetition" as if it were a single "participant" in the experiment. Therefore, we have 20 accuracy values, each averaged across 16 values from different ToM questions. The error bars then show the standard deviation of the 20 mean-accuracy values.

## Performance when Supported by Prompting

In-context learning via modified prompting boosted the ToM performance of all GPT models that were released after Davinci-2 (Figure 3).

First, we instructed the models to think step by step (SS). We found that SS thinking enhanced the performance of Davinci-3, GPT-3.5-Turbo, and GPT-4 (e.g. independent t-test for zero-shot vs SS thinking for GPT-4: $\Delta Acc$

= 0.10, p-value <0.001; zero-shot vs SS thinking for GPT-3.5-Turbo: ΔAcc = 0.06, p-value <0.001). However, SS thinking did not improve the accuracy of Davinci-2 (Figure 3).

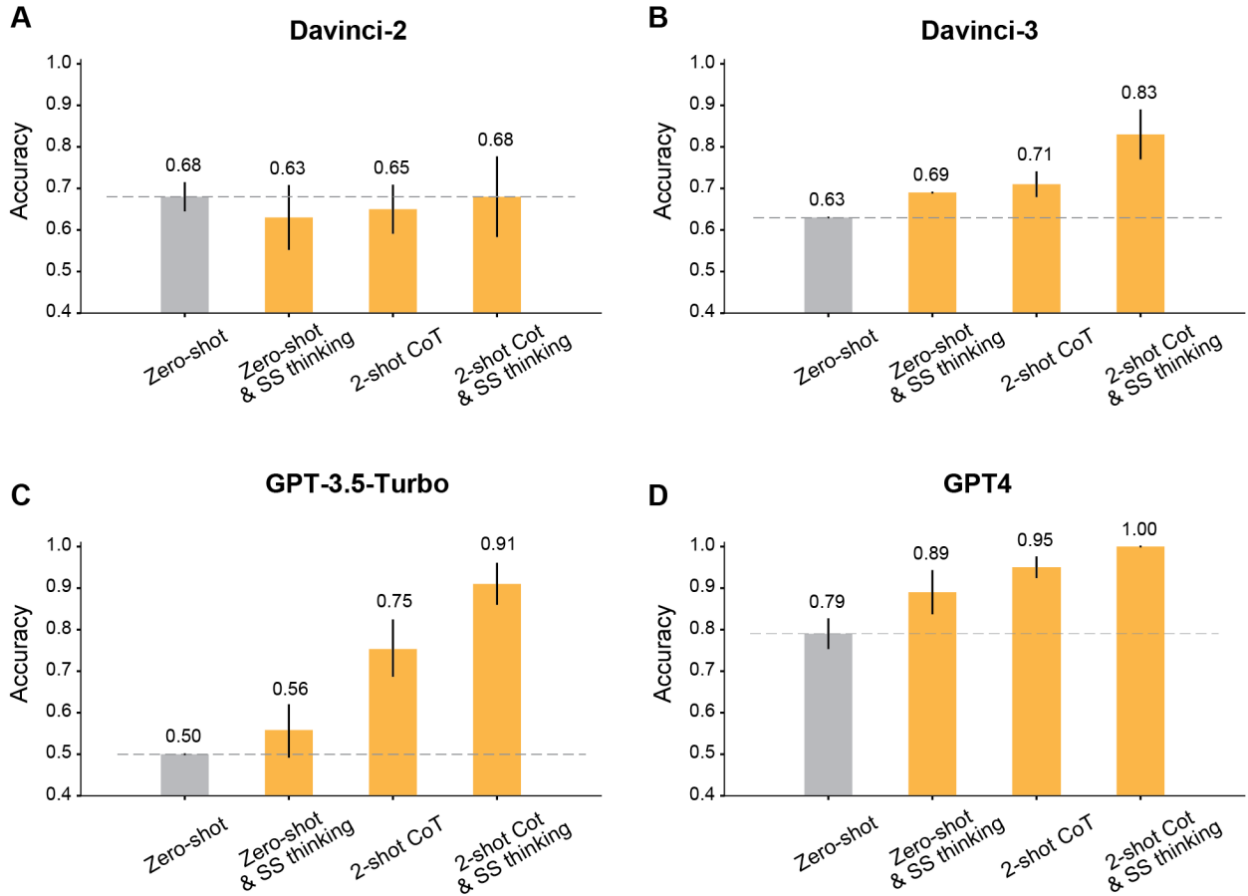

**Figure 3. Effects of In-context Learning Prompts on ToM performance in LLMs.** ToM performance of models using various in-context learning methods. For each model, the gray bar on the far left shows the Zero-Shot baseline ToM performance. The next three bars (orange) show the ToM performance on Zero-Shot plus SS Thinking; Two-Shot CoT; and Two-Shot CoT plus SS Thinking. Error bars indicate the standard deviation across 20 repetitions (see Figure 2, caption).

Second, we tested prompting with Two-shot chain-of-thought (CoT) reasoning. We found that Two-shot CoT increased the accuracy of all models that were trained with RLHF (all models except Davinci-2) (Figure 3). For GPT-3.5-Turbo, Two-shot CoT prompting significantly improved the performance above its zero-shot baseline (ΔAcc = 0.25, p-value <0.001), and was significantly more effective than prompting with SS thinking (ΔAcc = 0.19, p-value <0.001). For Davinci-3 and GPT-4, prompting with Two-shot CoT was slightly more effective than instructing step-by-step thinking.

Joint prompting with both Two-shot CoT reasoning and SS thinking produced the greatest increase in models' accuracy (Figure 3). ToM accuracy of all RLHF-trained models was significantly increased when the prompts included Two-shot CoT reasoning and SS thinking: Accuracy of Davinci-3 was increased by 20% ± 6% (mean

± std) relative to its zero-shot baseline (p-value <0.001). Accuracy of GPT-3.5-Turbo was increased by 41% ± 5% relative to its zero-shot performance (p-value <0.001). Finally, accuracy of GPT-4 increased by 21% to reach 100% accuracy (p-value <0.001, ceiling performance).

Altogether, appropriate prompting enabled all RLHF-trained models to achieve accuracy greater than 80%. When appropriately prompted, Davinci-3 achieved ToM accuracy of 83% (±6%), GPT-3.5-Turbo achieved 91% (±5%), and GPT-4 reached ceiling accuracy of 100%. Human performance in these scenarios was 87% (±4%) (See Supplement C).

**Interim Discussion**

*Do the increases in ToM performance arise from copying the reasoning steps from the prompt?*

The improved performance of LLMs via prompting does not appear to be due to simple mimicry of the specific reasoning steps provided in the chain-of-thought examples. The logic of reasoning in some scenarios was different from the reasoning logic of the 2 chain-of-thought examples. Both the in-context examples had the following essential reasoning logic: *Person P was not at location L when event E happened, so they are not aware of event E.* Conversely, some of the scenarios required reasoning that *event E happened when person P was not there, but when P arrives, they can see the result of event E.* If the improved performance was only due to copying a specific sort of reasoning it should not generalize across these distinct reasoning cases. To underscore this point, we performed the following analyses to test whether closely-related vs distantly-related CoT examples produce similar performance increases.

We hypothesized that if the improved ToM performance is due to copying the reasoning steps from the in-context ToM examples, then prompting with non-ToM examples should not enhance the ToM performance. To test this hypothesis, it is necessary to exclude the ToM questions for which the models consistently provided accurate zero-shot answers, because for such scenarios it is not possible to measure a performance increase. In other words, we focused on the ToM questions that the models could not correctly answer in zero-shot. Furthermore, we focused this analysis on Davinci-3 and GPT-3, because these models benefited from in-context chain-of-thought ToM examples and (in contrast to GPT-3.5-Turbo) they almost always returned a definitive response which did not require any subjective interpretation. Then for each model, we selected the scenarios that they could not correctly answer in zero-shot. This resulted in 4 scenarios for GPT-4 (mean zero-shot accuracy of 0.16), and 6 scenarios in Davinci-3 (zero-shot accuracy of 0.0). We then tested the ToM accuracy of the model for these selected scenarios under the following conditions: (i) Two-shot ToM CoT examples; (ii) Two-shot Non-ToM Inferential CoT examples; and (iii) Two-shot Photo CoT examples (Non-ToM Inferential examples; and Photo examples are shown in Supplement D). In the Non-ToM Inferential examples, questions were not about false belief or an agent's state of mind, but rather required inferential reasoning about the consequences of an event. Photo examples were selected from scenarios that described a scene or situation without an agent.

The Davinci-3 and GPT-4 models experienced increases in ToM performance from all of the classes of CoT examples that we tested: Photo examples, Non-ToM Inferential examples, and ToM examples. The mean accuracy increases for each model and each type of CoT example are shown in Figure 4, while the accuracy changes for individual ToM questions are shown in Figure S.1.

Prompting with Inferential and Photo examples boosted the models' performance on ToM scenarios even though these in-context examples did not follow the same reasoning pattern as the ToM scenarios. Therefore, our analysis suggests that the benefit of prompting for boosting ToM performance is not due to merely overfitting to the specific set of reasoning steps shown in the CoT examples. Instead, the CoT examples appear to invoke a mode of output that involves step-by-step reasoning, which improves the accuracy across a range of tasks.

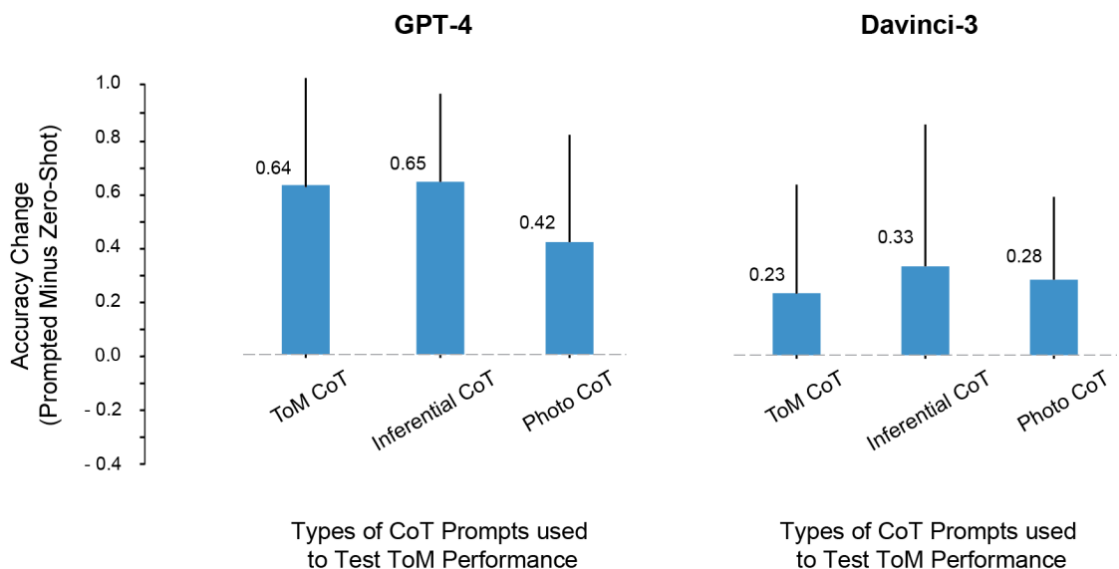

**Figure 4. Effects of Various Types of CoT Examples on ToM Performance.** The Y-axis represents the change in ToM accuracy compared to zero-shot (Two-shot CoT minus zero-shot), averaged across the scenarios that each model answered incorrectly during zero-shot testing (4 scenarios for GPT-4 and 6 scenarios for Davinci-3). The X-axis displays the types of in-context CoT examples provided to the model for evaluating ToM performance. The values indicate the mean change in accuracy compared to zero-shot for the assessed scenarios. For GPT-4, error bars are the standard deviation across the 4 accuracy-change values (1 value per scenario). For Davinci-3, error bars are the standard deviation across the 6 accuracy-change values. Refer to Figure S.1 for per-scenario accuracy results.

# General Discussion

We studied the performance of LLMs on comprehension tasks thought to require reasoning about an individual's state of mind. We evaluated the performance of Davinci-2, Davinci-3, GPT-3.5-Turbo, and GPT-4. When comprehension tasks required reasoning about superficially observable information in brief text scenarios (Photo scenarios), the zero-shot comprehension accuracy of LLMs increased monotonically from the oldest to newest LLMs. At the same time, the zero-shot accuracy on ToM questions did not consistently improve with the recency and sophistication of the models. However, with appropriate prompting, more recent models did exhibit consistent improvements in ToM performance. The prompt-improved performance approached or, in the case of GPT-4, exceeded human ToM performance on our test set.

Prompts that allowed for in-context learning boosted the expression of ToM reasoning in LLMs, relative to the zero-shot baseline. We used two recent prompting methods: step-by-step thinking (Kojima et al., 2022) and chain-of-thought reasoning (Wei et al., 2022). We found that all models, except Davinci-2, were able to exploit the modified prompting for higher ToM accuracy. The models exhibited the greatest increase in accuracy when the prompts combined both chain of thought reasoning and step-by-step thinking, rather than employing either alone. Also, the contrast between the performance of GPT-3.5-Trubo in zero-shot conditions and ICL conditions is a reminder that measured performance may not always reflect competence (Firestone, 2020). When provided with appropriate prompting, GPT-3.5-Turbo was capable of achieving higher performance than Davinci-2.

The GPT-3.5 variants that we tested (Davinci-2, Davinci-3, and GPT-3.5-Turbo) are similar in size (~175 B parameters) but differ in their training. In particular, Davinci-2 was the only model that was not finetuned with RLHF, and it was also the only model whose ToM performance was not increased by our prompt manipulations. It is possible that the RLHF component of the training enabled the models to exploit the in-context prompts in this setting.

LLMs may possess the capacity for performing ToM reasoning and yet not express this competence without the appropriate context or prompting. When supported by chain-of-thought and step-by-step prompting, Davinci-3 and GPT-3.5-Turbo exhibited ToM accuracy that was higher than GPT-4's zero-shot ToM accuracy. These results are not consistent with the claim that these models lack ToM reasoning capability (Sap et al., 2022), and they indicate the effectiveness of a prompting approach in enhancing LLM performance.

Models may fail for different reasons. For instance, in zero-shot inference, we noticed that Davinci-2 was more accurate than GPT-3.5 Turbo overall, but the models failed in different ways. When Davinci-2 answered a question incorrectly, it tended to do so with high confidence, without speculating over other possibilities. Conversely, GPT-3.5-Turbo's errors in zero-shot inference usually arose because the model's responses were more cautious: the model refrained from drawing confident conclusions and would frequently generate responses such as *there is not enough information in the scenario to answer this question*. This property may arise from GPT-3.5-Turbo's training method which was intentionally designed to make the model more careful. Generating inconclusive responses was also mentioned as one of the limitations of these models (OpenAI, 2022). Thus, the fact that Davinci-2 exhibited greater zero-shot ToM accuracy than GPT-3.5-Turbo is not conclusive evidence that Davinci-2 is more capable at ToM reasoning. In fact, compared to GPT-3.5-Turbo, Davinci-2 produced incorrect answers with higher confidence, made more error in linking one logical step to the next, and occasionally confabulated (a class of errors called *hallucinations* in the machine learning literature). Therefore, lower zero-shot performance of GPT-3.5-Turbo compared to Davinci-2 could be because GPT-3.5-Turbo has a stylistic bias that prevents it from providing definitive answers in short scenarios requiring inference.

Previous studies evaluating ToM performance in LLMs have primarily relied on single word completion or multiple-choice questions to measure their abilities (Kosinski, 2023; Sap et al., 2022; Trott et al., 2022; Ullman, 2023). However, this evaluation approach may not capture the sophistication of the ToM reasoning of which LLMs are capable. ToM reasoning is a complex behavior, which, even in humans, can involve multiple steps. Therefore, when responding to this challenging task, LLMs may benefit from producing longer-form answers. There are at least two reasons why LLMs may benefit in this way:

    First, we may be able to more fairly evaluate the model output when it is longer. LLMs sometimes generate the "correct" reasoning and then additionally mention other possibilities which lead it to arrive at an inconclusive overall summary. In such cases, the LLM is demonstrating that it is capable of correctly performing the reasoning steps for the ToM question, even though its overall conclusion does not correspond to one of a fixed set of options. Relatedly, the model might have a certain level of information regarding the potential results of a situation, but it may not be sufficient for it to draw a correct conclusion. This can be compared to the concept

that humans can still have some knowledge of an object's location despite having imperfect knowledge (Wu and Wolfe, 2018). Encouraging the model to systematically examine each piece of evidence and engage in a step-by-step reasoning process could help solidify its partial evidence and enable it to arrive at a definitive response.

Second, LLMs may unlock enhanced (or new) reasoning abilities when provided with the opportunity and the cues to elaborate a systematic step-by-step response. The improved ToM performance we observe is not simply a result of providing a number of in-context examples of ToM tasks (as in (Sap et al., 2022), where performance remained poor) but seems to rely on providing examples in which there is step-by-step inferential reasoning from the evidence before arriving at a conclusion (Figure 3 and Figure 4).

The LLMs may have seen some ToM or Photo scenarios during their training phase, but data leakage is unlikely to affect our findings. First, our findings concern the change in performance arising from prompting, and the specific prompts used to obtain this performance change were novel materials generated for this study. Second, if the model performance relied solely on prior exposure to the training data, there should be little difference between zero-shot Photo and ToM performance (Figure 2), as these materials were published in the same documents; however, the zero-shot performance patterns were very different across Photo and ToM scenarios. Third, the LLM performance improvements arose when the models elaborated their reasoning step-by-step, and this elaborated reasoning was not part of the training data. Therefore, although some data leakage is possible, it is unlikely to affect our conclusions concerning the benefits of prompting.

An important avenue for further testing is whether the prompt-driven performance gains are specific to ToM reasoning, or would be expected more generally in tasks involving other forms of inferential reasoning. Many of the ToM questions require the model to infer facts (e.g. mental states) that are not explicitly stated in the question, while (qualitatively speaking) it seems that many of the Control scenarios can be answered without performing as much inference beyond what is explicitly provided in the scenario text. Therefore, we are now testing LLMs comprehension in scenarios that require inferential reasoning but not reasoning about people's ToM. Our preliminary results indicate (i) a similar pattern in zero-shot performance for ToM scenarios and non-ToM scenarios which require inferential reasoning; and (ii) an improvement in non-ToM performance when incorporating the same prompts used for ToM scenarios (see Supplement E). Future research is needed to further explore the inferential reasoning capacity of LLMs as well as whether ToM inferences are a representative example of a more general set of inferential capabilities in LLMs.

We note four areas for improvement of this work. First, to evaluate ToM performance in our main analyses, we tested the effects of CoT prompting using only 2 CoT example scenarios, and we tested only 16 ToM questions, which were mostly probing agents' beliefs. Future research could explore the effects of different number of CoT examples, using various types of CoT examples, and examine a more diverse set of ToM tasks (Ullman, 2023). Second, in GPT-3.5 models, sometimes the reasoning was correct, but the model could not integrate that reasoning to draw the correct conclusion. Future research should extend the investigation of methods (such as RLHF) that can help LLMs draw a correct conclusion given the prior reasoning steps. Third, in the current study, we did not quantitatively analyze the failure modes of each model. To address the limitations of LLMs and further improve their reasoning capabilities, it is important to extend our understating of how and why different models fail. Moreover, we observed significant variability in performance across scenarios (Figure S.1). Therefore, measures of mean performance should be augmented by examination of failure modes and settings which may be specific to particular types of reasoning or subcomponents of ToM (Burnell et al., 2023). Finally, in the present study we manually scored the LLM responses. Because this form of labeling could be prone to individuals' interpretations, we are sharing the raw LLM outputs that were the basis of our findings (see Data Availability).[1] We await established benchmarks for evaluating complex reasoning and ToM behaviors in LLMs.

Our data do not speak to the question of whether LLMs possess a "mental faculty" that corresponds to a structured logical model of mental states. But our data do suggest that, when asking questions about ToM in LLMs, it will not be fruitful to seek a simple yes/no answer. The variation in performance across prompts may be analogous to how human cognition can vary across task contexts and motivational states, and how humans draw on more than one type of thinking (Evans, 2003). In LLMs, it is clear that task contexts (i.e. prompts) affect not only bottom-line accuracy, but, more qualitatively, the model's ability to invoke appropriate modes and styles of responding.

Our results are practically significant because they show how to aid LLMs in some forms of social reasoning. More abstractly, our results are another reminder that LLM behavior is highly complex and context sensitive. Therefore, it will be important to characterize their cognitive abilities via nuanced investigations (Firestone, 2020; Mitchell and Krakauer, 2022; Sejnowski, 2023), rather than reflexively applying existing cognitive ontologies. Also, as we build and interact with increasingly powerful cognitive artifacts, it is crucial to stretch our imaginations about what they are capable of and how they work.

# Conclusion

We have shown that LLMs can exploit chain-of-thought reasoning and step-by-step thinking to substantially improve their ToM performance. Human-level performance in these ToM scenarios was 87% ($\pm$ 4%). In contrast to zero-shot ToM settings, where only GPT-4 reached near 80% accuracy, with appropriate prompting, all RLHF-trained models exceeded 80% accuracy, with GPT-4 reaching ceiling accuracy (100%). Thus, appropriate prompting enhances the ToM reasoning performance of these highly context-sensitive models.

---

[1] Note that only a small portion of the data are edge cases which require subjective evaluation. A sample of edge case can be seen in Example 6 in the Correct Responses section. Furthermore, Davinci-2, Davinci-3, and GPT-4 were generally able to deliver decisive outcomes. Most of these edge cases primarily originated from GPT-3.5-Turbo, which was apparently finetuned to exercise caution in its responses.

# Acknowledgments

The authors gratefully acknowledge the support of the National Institutes of Mental Health (Grant R01MH119099). We further thank members of the Firestone Lab at Johns Hopkins University for helpful feedback on earlier drafts of this paper.

# Data Availability

The data used in this study are available at the following GitHub repository:

https://github.com/shrahimim/Boosting-Theory-of-Mind-in-LLMs-with-Prompting

# Supplementary Material:

# Supplement A

**16 Photo scenarios**

These scenarios are adapted and modified from (Dodell-Feder et al., 2011)**:**

1. "The traffic camera snapped an image of the black car as it sped through the stoplight. Soon after, the car was painted red and the license plates were changed." Q: Does the traffic camera show that the car is black?

2. "A map shows the ground floor plan. A photocopy was sent to the architect yesterday, but at the time the kitchen door was missing. It was added to the map this morning." Q: Does the architect's photocopy show the door?

3. "A photograph was taken of an apple hanging on a tree branch. The film took half an hour to develop. In the meantime, a strong wind blew the apple to the ground." Q: Does the developed photograph show the apple on the tree?

4. "To detect intruders, the lab uses an automated system for recording voices. In the empty lab one night, a computer error occurs and a synthetic voice reads the error message." Q: Is the number of people in the lab that night zero?

5. "A popular attraction in the park, pictured on many souvenirs, was a cliff face covered with ancient petroglyphs. Recently, the petroglyphs crumbled and scientists have not begun to restore them." Q: Today, can the petroglyphs be seen in the park?

6. "Sargent famously painted the south bank of the river in 1885. In 1910 a huge dam was built, flooding out the whole river basin, killing the old forests. Now the whole area is under water." Q: In the painting, is the south bank of the river wooded?

7. "When the picture was taken of the house, it was one story tall. Since then, the renovators added an additional story and a garage." Q: Is the house currently one story?

8. "Accounts of the country's economic success were recorded in books from the early 1900s. Soon after, a horrible plague hit the country, and the country was sent into an economic depression." Q: Do early 1900s novels portray the country as experiencing economic wealth?

9. "Part of the garden is supposed to be reserved for the roses; it's labeled accordingly. Recently the garden has run wild, and dandelions have taken over the entire flower bed." Q: Does the label say these flowers are roses?

10. "A long time ago, an explorer mapped a small island. Since then, the water levels rose and only a tiny part of the island is now left above water." Q: On the explorer's maps, does the island appear to be mostly above water?

11. "A large oak tree stood in front of City Hall from the time the building was built. Last year the tree fell down and was replaced by a stone fountain." Q: Does an antique drawing of City Hall show a fountain in front?

12. "A volcano erupted on a Caribbean island three months ago. Barren lava rock is all that remains today. Satellite photographs show the island as it was before the eruption." Q: Do satellite photographs show the island is covered in lava?

13. "The family's old video tape recorded the daughter's first birthday party at their house in Chicago. Since then, the family sold their house and moved to San Francisco." Q: Does the video show the family living in San Francisco?

14. "At the time a portrait was drawn of a young man, he had short brown hair and no facial hair. Now the man's hair is long and gray and so is his beard." Q: Today, is the length of the man's beard long?

15. "The girl's middle school pictures showed her wearing a white blouse. Later, a red sock was accidentally washed with the blouse and the blouse turned pink." Q: Was the color of the blouse in the pictures pink?

16. "A small leaf was placed on a wet clay flower pot. When the pot was baked at high temperatures to harden the clay, the leaf crumbled, but its impression remained." Q: Is the actual leaf intact?

# Supplement B

## 16 ToM scenarios & the scenarios used for in-context learning:

16 ToM scenarios are adapted and modified from (Dodell-Feder et al., 2011).

1. "The weather was so warm today that all the tulips in Pam's backyard suddenly bloomed. The tulips next to Pam's office still have not yet flowered, though. Pam has been at work all day." Q: When Pam is driving home after work, does she assume her tulips have bloomed? A:

2. "Susie parked her sports car in the driveway. In the middle of the night, Nathan moved her car into the garage to make room for his minivan. Susie woke up early in the morning." Q: When Susie wakes up, does she see the minivan in the driveway? A:

3. "The morning of the high school dance Sarah placed her high heel shoes under her dress and then went shopping. That afternoon, her sister borrowed the shoes and later put them under Sarah's bed." Q: When Sarah gets ready, does she assume her shoes are under her dress? A:

4. "Sally and Greg called ahead of time to make a reservation for the back-country cabin. The park ranger forgot to write down the reservation and two other hikers got to the cabin first." Q: When Sally and Greg arrive, do they find their cabin unoccupied? A:

5. "When Jeff got ready this morning, he put on a light pink shirt instead of a white one. Jeff is colorblind, so he can't tell the difference between subtle shades of color." Q: Does Jeff believe he is wearing a white shirt? A:

6. "Larry chose a debated topic for his class paper due on Friday. The news on Thursday indicated that the debate had been solved, but Larry never read it." Q: When Larry writes his paper, does he think the debate has been solved? A:

7. "Every day Jill goes to the coffee shop on the corner and orders a latte, her favorite drink. Today, the cashier misunderstands Jill and prepares a mocha instead." Q: Does Jill think her drink will taste like a mocha? A:

8. "Expecting the game to be postponed because of the rain, the Garcia family took the subway home. The score was tied, 3-3. During their commute the rain stopped, and the game soon ended with a score of 5-3." Q: When the Garcia family arrives home, do they believe the score is 5-3? A:

9. "Amy walked to work today. When George woke up, he saw her car in the drive. Her room was quiet and dark. George knows that when Amy is sick, she lies down in a dark room." Q: Does George think Amy is sick today? A:

10. "John told Mary that he had lost his keys. The two of them searched the house with no luck. Then Mary went outside to look in the car. Suddenly John noticed his keys behind the sofa." Q: By the time Mary comes in, does John know where his keys are? A:

11. "At night a bear broke into a cooler near a tent and drank the soda. Five hours later, the campers woke up and went to their cooler for breakfast." Q: Do the campers find their cooler empty of soda? A:

12. "When Lisa left Jacob, he was deep asleep on the beach. A few minutes later a wave woke him. Seeing Lisa was gone, Jacob decided to go swimming." Q: Does Lisa now believe that Jacob is asleep? A:

13. "Jenny put her chocolate away in the cupboard. Then she went outside. Alan moved the chocolate from the cupboard into the fridge. Half an hour later, Jenny came back inside." Q: Does Jenny expect to find her chocolate in the cupboard? A:

14. "A window wiper was commissioned by a CEO to wipe an entire building. He finished the right side, but his platform broke before he could do the left side. The next morning the CEO arrived with foreign investors." Q: When the CEO comes to work, do they discover that all of the windows are cleaned? A:

15. "Hopeful to catch a prize fish, George went fishing. That afternoon, he saw his fishing line bend over as if he had caught a big fish. Actually, George's fishing pole had snagged a small tire." Q: At the end of the fishing line, does George see a fish? A:

16. "Laura didn't have time to braid her horse's mane before going to camp. While she was at camp, William brushed Laura's horse and braided the horse's mane for her." Q: When Laura is at camp, does she assume that her horse's hair is braided? A:

## The scenarios used in the chain-of-thought reasoning examples:

### Example 1:
"Anne made lasagna in the blue dish. After Anne left, Ian came home and ate the lasagna. Then he filled the blue dish with spaghetti and replaced it in the fridge."
Q: Does Anne think the blue dish contains spaghetti?
A: When Anne left, the blue dish contained lasagna. Ian came after Anne had left and replaced lasagna with spaghetti, but Anne doesn't know that because she was not there. So, the answer is: No, she doesn't think the blue dish contains spaghetti.

### Example 2:
"The girls left ice cream in the freezer before they went to sleep. Over night the power to the kitchen was cut and the ice cream melted."
Q: When they get up, do the girls believe the ice cream is melted?
A: The girls put the ice cream in the freezer and went to sleep. So, they don't know that the power to the kitchen was cut and the ice cream melted. So, the answer is: No, the girls don't believe the ice cream is melted.

# Supplement C

## ToM accuracy of human participants

To measure humans' performance in ToM and Photo scenarios, we recruited 125 online participants through the Qualtrics platform. Participants were 18 to 65 years old, native English speakers, and located in the United States.

Participants had 18 seconds to read each scenario. Once the 18-second duration was over, they were immediately directed to a new screen showing the comprehension question. They could respond to the questions at their own pace by clicking on one of the two options (Yes/No). The Photo scenarios and ToM scenarios were interleaved and randomized such that each participant experienced both types of scenarios.

The Photo and ToM scenarios were of comparable difficulty for human participants. Participants showed similar accuracy across Photo (86% ±4%) and ToM (87% ±4%) scenarios.

From these data we conclude that:

i) the Photo scenarios are a good control condition for the ToM scenarios, because of their matched performance in humans.
ii) the scenarios are relatively easy for humans to solve, but not so easy that humans are at ceiling performance (given the modest incentives for accuracy in online participants).

Making a direct quantitative comparison between human and LLM performance is not warranted, because of the many differences in the testing conditions. However, the human performance does indicate that the questions are sufficiently difficult that humans occasionally (>10% of the time) make errors, perhaps because they overlook details in the scenario or because they make unusual assumptions or inferences.

# Supplement D

## Effects of Various In-context Examples on ToM Performance of LLMs

We tested the efficacy of Photo examples, Non-ToM Inferential examples, and ToM examples in enhancing ToM performance when providing chain-of-thought prompts. We used the following examples in each class:

1) **Photo Chain-of-Thought Examples:**

Example 1:
> Scenario: "A long time ago, an explorer mapped a small island. Since then, the water levels rose and only a tiny part of the island is now left above water."
> Q: On the explorer's maps, does the island appear to be mostly above water?
> A: An explorer mapped an island a long time ago. Then water levels rose. Now only a tiny part of the island is above water. The maps show the island before water levels rose. Therefore, in the map most of the island is above water. So, the answer is: Yes, in the maps, most of the island is above the water.

Example 2:
> Scenario: "A volcano erupted on a Caribbean island three months ago. Barren lava rock is all that remains today. Satellite photographs show the island as it was before the eruption."
> Q: Do satellite photographs show the island is covered in lava?
> A: A volcano erupted three months ago and as a result lava rock is all that remains today. Since the photograph shows the island before the eruptions, it does not show the island covered in Laval. So, the answer is: No, the photographs does not show the island covered in lava.

2) **Non-ToM Inferential Chain-of-Thought Examples:**

Example 1:
> Scenario: "Sarah was excited to show off her new dress at the party. As she walked in, she noticed that another girl was wearing the exact same dress. She quickly went back home to change and then returned to the party."
> Q: Did Sarah wear her favorite dress the whole night?
> A: Sarah was excited about her new dress and wanted to wear it as the party. But since another girl was wearing the same dress, she went home and changed to a different dress. So, the answer is: No, she did not wear her favorite dress the whole night.

Example 2:
> Scenario: "Lucy was in a hurry to get to work, so she quickly grabbed her lunch from the fridge and left the house. At lunchtime, she opened her lunchbox and realized she had accidentally taken her roommate's lunch."
> Q: Will Lucy's roommate have her own lunch to eat that day?
> A: Lucy was in hurry and mistakenly grabbed her roommate lunch. So, her roommate's lunch is now with Lucy. So, the answer is: No, Lucy's roommate will not have her own lunch to eat that day.

3) **ToM Chain-of-Thought Examples:**

Example 1:
> Scenario: "Anne made lasagna in the blue dish. After Anne left, Ian came home and ate the lasagna. Then he filled the blue dish with spaghetti and replaced it in the fridge."
> Q: Does Anne think the blue dish contains spaghetti?

A: When Anne left the blue dish contained lasagna. Ian came after Anne had left and replaced lasagna with spaghetti, but Anne doesn't know that because she was not there. So, the answer is: No, she doesn't think the blue dish contains spaghetti.

Example 2:
> Scenario: "The girls left ice cream in the freezer before they went to sleep. Over night the power to the kitchen was cut and the ice cream melted."
> Q: When they get up, do the girls believe the ice cream is melted?
> A: The girls put the ice cream in the freezer and went to sleep. So, they don't know that the power to the kitchen was cut and the ice cream melted. So, the answer is: No, the girls don't believe the ice cream is melted.

## **GPT-4**

For this analysis, the performance of GPT-4 was assessed on the 4 scenarios below. These are scenarios that GPT-4 answered incorrectly in zero-shot testing.

**Scenario 1:**
> "Amy walked to work today. When George woke up, he saw her car in the drive. Her room was quiet and dark. George knows that when Amy is sick, she lies down in a dark room."
> Q: Does George think Amy is sick today?

**Scenario 2:**
> "When Lisa left Jacob, he was deep asleep on the beach. A few minutes later a wave woke him. Seeing Lisa was gone, Jacob decided to go swimming."
> Q: Does Lisa now believe that Jacob is asleep?

**Scenario 3:**
> "The weather was so warm today that all the tulips in Pam's backyard suddenly bloomed. The tulips next to Pam's office still have not yet flowered, though. Pam has been at work all day."
> Q: When Pam is driving home after work, does she assume her tulips have bloomed?

**Scenario 4:**
> "Every day Jill goes to the coffee shop on the corner and orders a latte, her favorite drink. Today, the cashier misunderstands Jill and prepares a mocha instead."
> Q: Does Jill think her drink will taste like a mocha?

## **Davinci-3**

For this analysis, the performance of Davinci-3 was assessed on the 6 scenarios below. These are scenarios that Davinci-3 answered incorrectly in zero-shot testing.

**Scenario 1)**
> "Susie parked her sports car in the driveway. In the middle of the night, Nathan moved her car into the garage to make room for his minivan. Susie woke up early in the morning."

      Q: When Susie wakes up, does she see the minivan in the driveway?

**Scenario 2)**
      "When Jeff got ready this morning, he put on a light pink shirt instead of a white one. Jeff is colorblind, so he can't tell the difference between subtle shades of color."
      Q: Does Jeff believe he is wearing a white shirt?

**Scenario 3)**
      "When Lisa left Jacob, he was deep asleep on the beach. A few minutes later a wave woke him. Seeing Lisa was gone, Jacob decided to go swimming."
      Q: Does Lisa now believe that Jacob is asleep?

**Scenario 4)**
      "Laura didn't have time to braid her horse's mane before going to camp. While she was at camp, William brushed Laura's horse and braided the horse's mane for her."
      Q: When Laura is at camp, does she assume that her horse's hair is braided?

**Scenario 5)**
      "The morning of the high school dance Sarah placed her high heel shoes under her dress and then went shopping. That afternoon, her sister borrowed the shoes and later put them under Sarah's bed."
      Q: When Sarah gets ready, does she assume her shoes are under her dress?

**Scenario 6)**
      "Jenny put her chocolate away in the cupboard. Then she went outside. Alan moved the chocolate from the cupboard into the fridge. Half an hour later, Jenny came back inside."
      Q: Does Jenny expect to find her chocolate in the cupboard?

Per-scenario accuracies are shown in Figure S.1. The mean accuracy for this analysis is shown in Figure 4.

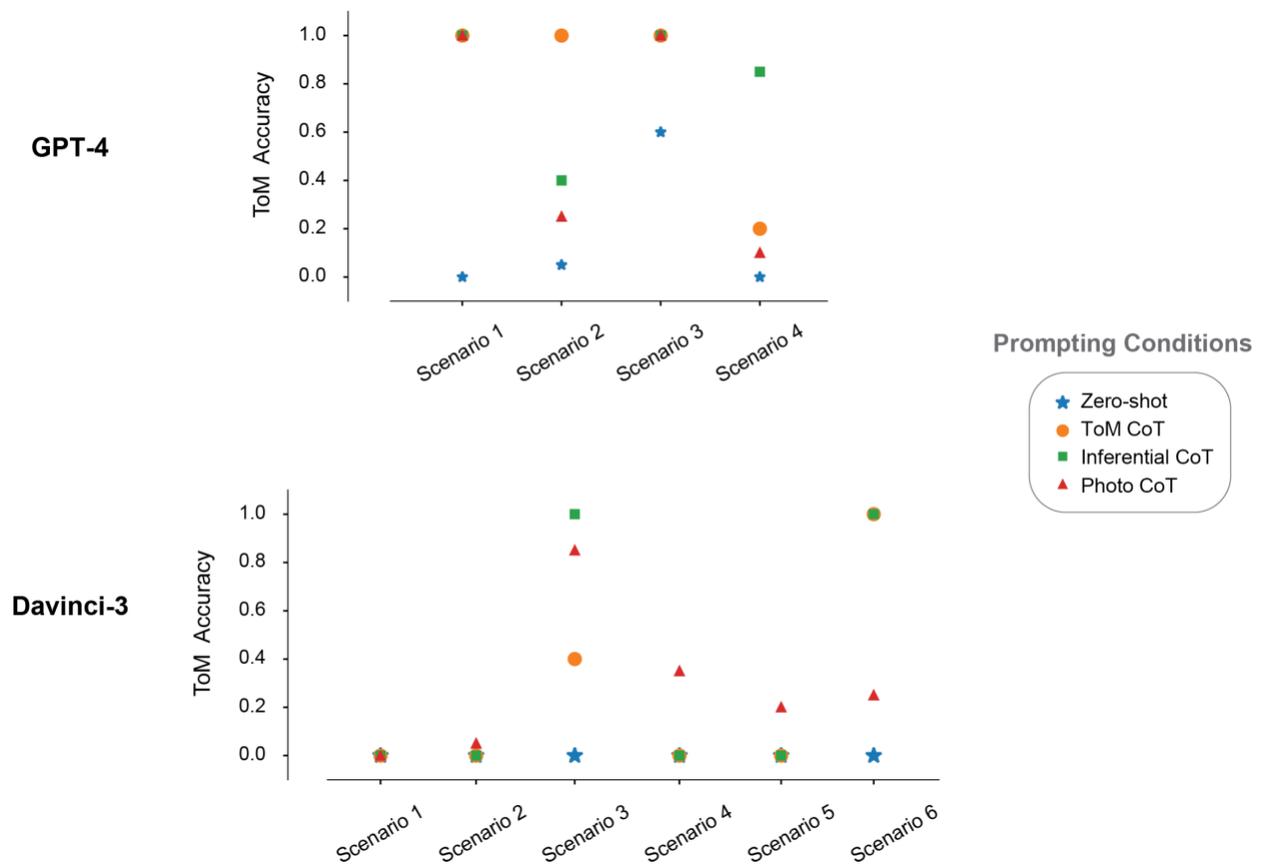

**Figure S.1. Per-Scenario Accuracy using 4 prompting conditions in GPT-4 and Davinci-3.** For each condition (e.g. Inferential CoT), the per-scenario accuracy is calculated based on the mean accuracy across 20 repetitions of a single prompt.

# Supplement E

## Generalizability of ToM Results to Inferential Reasoning in Non-ToM Scenarios

To investigate the generalizability of our prompting methods, we designed and tested a set of scenarios that required inferential reasoning, but which were not false-belief scenarios. In these new scenarios, the reasoning does not follow the same steps as in the false-belief scenarios (e.g. *person P does not know about event E,* etc). Therefore, the specific reasoning steps used in the in-context ToM CoT examples cannot be directly transferred to these non-ToM scenarios. We created and tested the following non-ToM scenarios:

1. "Anna was baking cookies for her friends. She realized she was out of sugar and went to the store to buy some. While she was gone, her dog, Max, jumped on the counter and ate the cookie dough." Q: Will Anna be able to finish making cookies after returning?

2. "Lisa was running late for work, so she quickly grabbed her lunch out of the fridge and left the house. Seeing Lisa leaving the house, her cat knocked over her coffee mug on the kitchen counter." Q: Will Lisa find the kitchen counter clean when she returns?

3. "Lucy was in a hurry to get to work, so she quickly grabbed her lunch from the fridge and left the house. At lunchtime, she opened her lunchbox and realized she had accidentally taken her roommate's lunch." Q: Will Lucy's roommate have her own lunch to eat that day?

4. "Lucy was baking a cake for her friend's birthday. She put the cake in the oven and set a 20-minute timer. She then went to the living room to watch her favorite show for 30 minutes." Q: Will the cake be overcooked by the time Lucy checks the oven?

5. "Tim and Amy went on a hike through the forest. After walking for several hours, they came across a bear. They quickly and quietly backed away, retracing their steps back to the starting point." Q: Did Tim and Amy finish their hike?

6. "Sarah was excited to show off her new dress at the party. As she walked in, she noticed that another girl was wearing the exact same dress. She quickly went back home to change and then returned to the party." Q: Did Sarah wear her favorite dress the whole night?

7. "During a thunderstorm, Sarah closed all the windows in her house. She then went to the basement to do laundry. While she was downstairs, a tree branch fell and broke one of the windows upstairs." Q: When Sarah finishes the laundry, does she find all the windows closed?

8. "Lucy spent hours preparing a delicious meal for her husband's surprise birthday party. Just as she was setting the table, she accidentally knocked over a glass of red wine, spilling it all over the food. She ordered pizza as a last-minute replacement." Q: Does Lucy serve her homemade meal to the guests?

We tested two conditions: (1) The zero-shot performance of the LLMs in these new scenarios; (2) The efficacy of the same prompts, Two-shot chain-of-thought ToM examples plus step-by-step thinking, for these new set of non-ToM scenarios.

We found a similar pattern between the performance in the non-ToM scenarios and the ToM scenarios (Figure S.2): First, similar to results from Figure 2B, the zero-shot accuracy dropped from Davinci-2 to Davinci-3, and from Davinci-3 to GPT-3.5-Turbo[2], while GPT-4 showed the highest accuracy, reaching ceiling performance. Second, similar to the results from Figure 3, we found that prompting significantly boosted the ToM accuracy for Davinci-3 and GPT-3.5-Turbo, while it did not improve the accuracy of Davinci-2. The performance of GPT-4 was at ceiling in the zero-shot setting, so there was no room for benefit from prompting. Thus, this preliminary analysis indicates that the performance gains from prompting are not limited to inferential ToM tasks, but may extend to other inferential tasks.

This result is notable for two reasons: (i) The reasoning steps for the chain-of-thought examples (which were ToM examples) were different from the reasoning steps required to answer the questions (which were non-ToM examples). Therefore, these performance increases are not merely due to copying the specific reasoning steps in demonstrated in the prompt. (ii) These questions do not ask about an agent's state of the mind, but instead require inference about a particular situation. Therefore, it is possible that our primary findings concerning ToM may generalize to a broader class of inferential reasoning tasks.

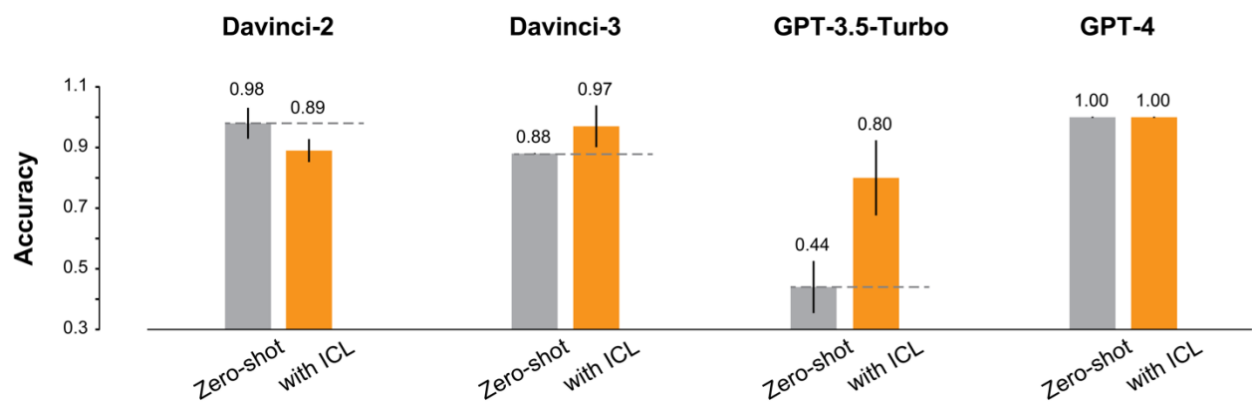

**Figure S.2. Accuracy of LLMs in Non-ToM Scenarios.** Accuracy of Davinci-2, Davinci-3, GPT-3.5-Turbo, and GPT-4 in new scenarios that do not involve reasoning about agents' beliefs. The gray bars (left bars) show the zero-shot performance. The orange bars (right bars) show the performance with prompts that allow for in-context learning (ICL). The ICL prompt includes both Two-shot CoT plus instructing step-by-step thinking. The accuracy is the mean of 160 values (8 scenarios x 20 repetitions each). The error bars show the standard deviation of the accuracy over 8 scenarios for each of 20 repetitions (standard deviation across 20 values, each a mean of 8 scenarios).

---

[2] Data from GPT-3.5-Turbo for this particular analysis was collected before it was possible to control the temperature parameter at inference time using the OpenAI API. Therefore, for this particular analysis, the temperature parameter for GPT-3.5-Turbo was the default value (likely 0.7), whereas for the other three models the temperature was set to 0.4.